\documentclass[runningheads]{llncs}
\usepackage{graphicx}

\usepackage{tikz}
\usepackage{comment}
\usepackage{amsmath,amssymb} 
\usepackage{color}

\usepackage{booktabs}

\usepackage[accsupp]{axessibility}  

\usepackage{multirow}

\begin{document}
\pagestyle{headings}
\mainmatter
\def\ECCVSubNumber{12}  

\title{Model-Assisted Labeling via Explainability for Visual Inspection of Civil Infrastructures}

\titlerunning{ECCV-22 submission ID \ECCVSubNumber} 
\authorrunning{ECCV-22 submission ID \ECCVSubNumber} 
\author{Anonymous ECCV submission}
\institute{Paper ID \ECCVSubNumber}

\titlerunning{Model-Assisted Labeling via Explainability}
%

\author{Klara Janouskova\inst{1} \thanks{Part of the work was done during the author's internship at IBM Research Zürich} \and
Mattia Rigotti\inst{2} \and
Ioana Giurgiu\inst{2} \and Cristiano Malossi\inst{2}}
%
\authorrunning{K. Janouskova et al.}
%
\institute{Visual Recognition Group, Faculty of Electrical Engineering, Czech Technical University in Prague \\
\email{klara.janouskova@fel.cvut.cz}\\
\and
IBM Research Zürich \\
\email{\{mrg, igi, acm\}@zurich.ibm.com}}
\maketitle

\begin{abstract}
Labeling images for visual segmentation is a time-consuming task which can be costly, particularly in application domains where labels have to be provided by specialized expert annotators, such as civil engineering.
In this paper, we propose to use attribution methods to harness the valuable interactions between expert annotators and the data to be annotated in the case of defect segmentation for visual inspection of civil infrastructures.
Concretely, a classifier is trained to detect defects and coupled with an attribution-based method and adversarial climbing to generate and refine segmentation masks corresponding to the classification outputs. These are used within an assisted labeling framework where the annotators can interact with them as proposal segmentation masks by deciding to accept, reject or modify them, and interactions are logged as weak labels to further refine the classifier.
Applied on a real-world dataset resulting from the automated visual inspection of bridges, our proposed method is able to save more than 50\% of annotators' time when compared to manual annotation of defects.

\keywords{civil infrastructure, weakly supervised learning, semantic segmentation, model-assisted labeling}
\end{abstract}

\section{Introduction}

Until  recently  visual  inspection  was  exclusively  a  manual process  conducted  by  reliability  engineers.  Not  only  is  this dangerous due to the complexity of many civil engineering structures and the fact that some parts are hardly accessible. The  main  objective of  the  inspection  is  to  assess  the  condition  of  an  asset  and determine  whether  repair  or  further  maintenance  operations are  needed.  Specifically,  engineers  make  such  decisions  by analyzing  the  surfaces in search for defects,  such  as  cracks, spalling, rust or algae, and assessing their severity, relative to their size and location in the structure. 

The advances in drone technology and its falling costs have recently  pushed  this  laborious  process  of  manual  inspection progressively  towards  automation.  Flying  drones  around  a structure and   using   embedded   high-resolution   cameras   to collect  visual  data  from  all  angles  not  only  speeds  up  the inspection   process,   but   it   also   removes   the   human   from potentially  dangerous  situations.  In  addition,  thanks  to  the power  of  artificial  intelligence  capabilities,  defects  can  be detected  and  localized  with  high  precision  automatically  and presented  to  the  reliability  engineer  for  further  analysis.

Typical approaches go beyond defect detection and generate fine-grained segmentation masks, which better characterize the defect.
However, the drawback of these segmentation models is that they are fully supervised and therefore require a significant volume of high quality annotations at training time.
Generating fine-grained segmentation masks is a manual task that involves a human expert deciding whether each pixel in the image belongs to a defect or not, which is time consuming and error-prone.
Depending on the size of the images captured during inspection and the volume of defects present in a single image, annotating all defects per image can take hours, even with the aid of annotation tools like CVAT~\cite{cvat} or SuperAnnotate~\cite{superann}. For example, it has been reported that single large (2048x1024) images depicting complex scenes require more than 90 minutes for pixel-level annotation~\cite{cityscapes}. 

The need for such expensive annotations can be alleviated by weakly supervised learning, in which a neural network is trained with cheaper annotations than explicit localization labels. In particular, weakly supervised segmentation methods can use image-level class labels~\cite{kim,lee,ahn,chang}, which require a single pixel annotation within the localized region of the target object. By using attribution maps obtained from a classifier, such as Grad-CAM~\cite{selvaraju2017grad}, it is possible to identify the most important and discriminative regions of an image. However, these generated maps do not tend to cover the entire region of the target objects. Typical attempts to extend the maps manipulate either the image~\cite{wheretolook,hideandseek}, or the feature map~\cite{hou,zhang}.

\begin{figure}[t]
    \centering
    \includegraphics[keepaspectratio, width=0.8\linewidth]{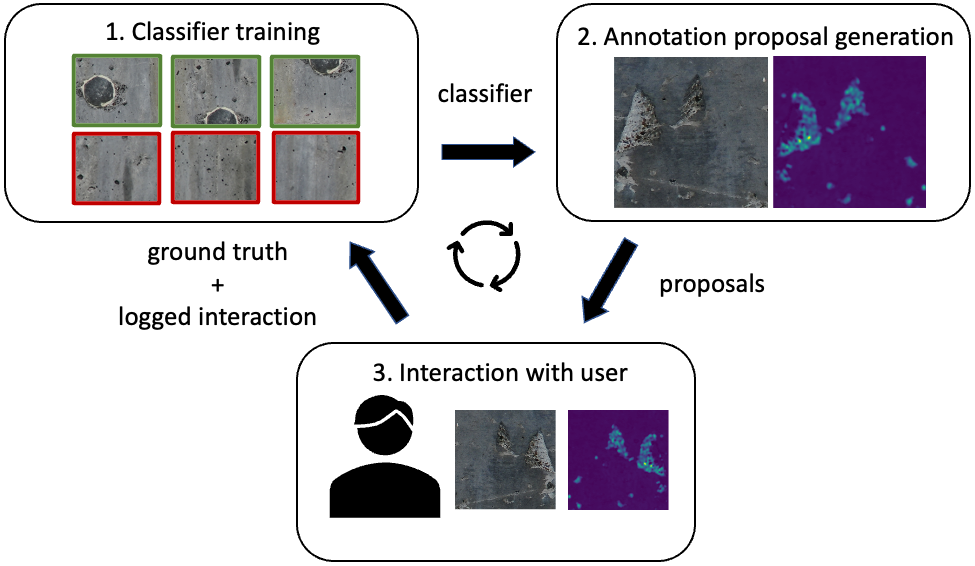}
    \caption{Overview of the assisted labeling framework. First, a classifier is trained on weakly annotated images. Second, the classifier generates annotation proposals. Finally, the user interacts with the proposal (accept/modify/reject). The interaction is logged and used to extend the training set and improve the classifier, resulting in improved annotation proposals on future data. }
    \label{fig:framework_overview}
\end{figure}

In this paper, we employ a different approach, based on adversarial-climbing, to extend the attributed regions of a target object~\cite{advcam}. This is opposed to an adversarial attack, which generates small perturbations of an image in order to change its classification output. As a result of applying adversarial climbing iteratively, the attribution map of the image gradually focuses on more extended regions of the target object, and can be used to generate fine-grained segmentation masks.

Specifically, we build a framework for model-assisted labeling of defects detected as a result of visual inspections of bridge structures from high resolution images. We train a classifier to recognize defect labels, apply Grad-CAM to generate segmentation masks and refine these masks with adversarial climbing. Once the masks have been generated, they are made available to the user through an interaction tool, where the expert is able to visualize, accept, reject or correct them, if need be (Figure~\ref{fig:framework_overview}). We evaluate the  approach on a real-world dataset and show that even after the first iteration, more than 50\% of annotators' time is saved by refining the obtained masks instead of manually generating them. Moreover, the time saved is expected to increase in further iterations.

\section{Related Work}

\subsection{Weakly supervised segmentation and localization}
The vast majority of weakly supervised semantic segmentation and object localization methods depend on attribution maps obtained with approaches like Grad-CAM~\cite{selvaraju2017grad} from
a trained classifier. While identifying the relevant regions of an image that have contributed to the classifier's decision is the goal, these regions tend to not be able to identify the whole region occupied by the target object.

Therefore, there have been many attempts to extend the attributed regions to ensure they cover more of the detected object. One popular approach is to manipulate the image~\cite{wheretolook,hideandseek}. For instance through erasure techniques~\cite{erasure1,hou,zhang,erasure4,erasure5}, already identified discriminative regions of the image are removed in an iterative manner, thus forcing the classifier to identify new regions of the object to be detected. However, the main drawback of the erasure approach is that there is a risk to generate wrong attribution maps when by erasing the discriminative regions of an image, the classifier's decision boundary changes. An alternative to image manipulation is feature map manipulation~\cite{fickle,tpami}. This produces a unified feature map by aggregating a variety of attribution maps from an image obtained by applying dropout to the network's feature maps.

Recently, adversarial climbing has been proposed to extend the attributed regions of the target object~\cite{advcam}. Applied iteratively to the attribution maps of a manipulated image results in a gradual identification of more relevant regions of the object. Regularization is additionally applied to avoid or reduce the activation of irrelevant regions, such as background or regions of other objects. Unlike other approaches that require additional modules or different training techniques, applying adversarial climbing acts essentially as a post-processing step on top of the trained classifier. This makes it possible and easy to replace the underlying classifier's architecture or improve its performance without performing any changes to the backbone.

While adversarial climbing has been mainly applied for semantic segmentation, we employ it for instance segmentation, to generate precise and high-quality segmentation masks for fine-grained defects present in civil infrastructures. These masks go beyond providing localization cues for weakly supervised instance segmentation and defect localization. They significantly reduce the time required to manually annotate such defects at pixel-level, thus enabling downstream tasks such as supervised defect detection and segmentation at much lower costs.  

\subsection{Annotation tools}
Many annotation tools successfully deploy semi-supervised interactive annotation models, with different level of weak supervision at inference time. Traditional methods like GrabCut~\cite{grabcut} which do not require fully-supervised pre-training exist, but are outperformed by learning-based strategies.

In DEXTR~\cite{dextr}, at least four extreme points are required at inference time to infer segmentation while a bounding box and up to four correction points are used in~\cite{largeintseg}. In ~\cite{nuclick}, a single click on an instance is enough to generate its segmentation mask.
A crucial disadvantage of these approaches is that they do not have any localization ability and the detection of the defects fully relies on the human annotator.
The performance of learning-based models also typically decreases with domain transfer.
To obtain good performance on a new domain, full annotations are required.
In~\cite{efficientintseg}, the problem of domain transfer is tackled by online fine-tuning.

\section{Model-Assisted Labeling Framework}

Instead of requiring segmentation masks from annotators, we propose to use weak labels consisting of classification labels. One label per image would make GPU training extremely challenging due to the large size of images in our dataset. Therefore, we ask the annotators to localize patches that contain defects. This approach is still substantially faster to input since they only require one click per defect.
Similarly, negative samples require one click to indicate the absence of defects within a given image patch.
These inputs are then used to generate a training dataset for a defect classifier by sampling crops around the annotated pixels.

Common approaches to weakly supervised learning with class-level supervision~\cite{weak_irn} use encoder architectures such as ResNet~\cite{resnet} to generate class activation maps (CAMs) ~\cite{selvaraju2017grad}. 

Some work has gone into improving the resolution of the obtained CAMs using multi-scale inference~\cite{weak_irn,advcam} followed by post-processing steps like dense CRF~\cite{dense_crf}, or aggregating activations from different levels of a ConvNet~\cite{layercam,polycam,hrnet}.
In a semi-supervised setup, \cite{cam_unet} adopt a U-net architecture~\cite{unet} and pre-train the encoder on a classification task, and then train the decoder to improve the mask starting with CAMs as a segmentation prior.

\subsection{Proposal generation and refinement}

\begin{figure*}[htb]
    \centering
        \begin{tabular}{ccc}
             \includegraphics[keepaspectratio, width=0.28\linewidth]{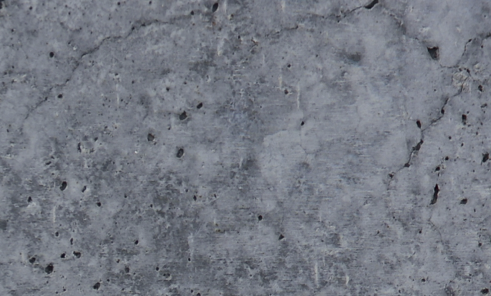} & 
             \includegraphics[keepaspectratio, width=0.28\linewidth]{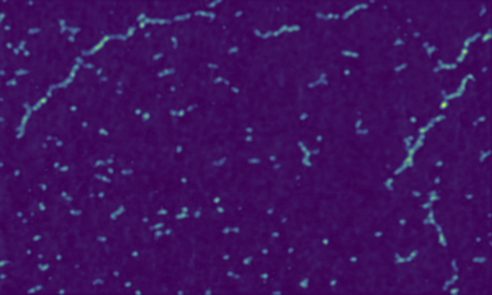} &
             \includegraphics[keepaspectratio, width=0.28\linewidth]{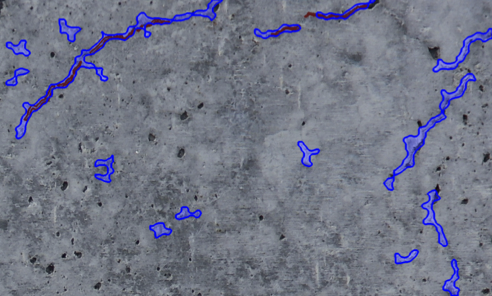} \\
             
             \includegraphics[keepaspectratio, width=0.28\linewidth]{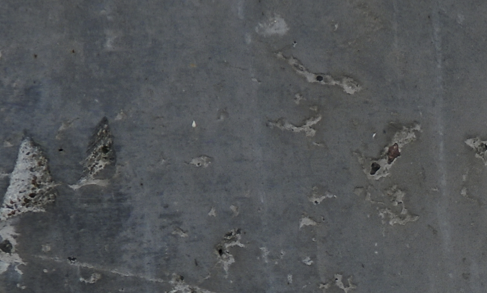} & 
             \includegraphics[keepaspectratio, width=0.28\linewidth]{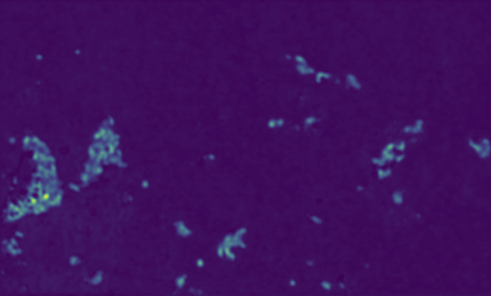} &
             \includegraphics[keepaspectratio, width=0.28\linewidth]{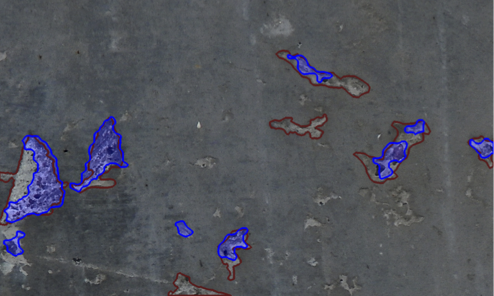} \\
             
             Original images & Class Activation Maps & Proposed (blue) and\\ & & GT (red) segmentations \\
        \end{tabular}
    \caption{Example images for the class `crack'. First column: two examples of image of concrete surfaces containing cracks, and therefore labeled with the classification label `crack'. Middle column: corresponding CAMs. Last column: corresponding proposed annotations obtained by filtering out noise in the CAMs by post-processing (blue), Ground Truth (GT) segmentation masks obtained by expert manual annotation from scratch (red). The proposed annotations generated by our method have large overlap with GT segmentation masks provided by expert annotators.}
    \label{fig:gt_overlay_examples}
\end{figure*}

Our weakly-supervised method to generate fine-grained segmentation masks consists of two steps. First, a deep neural network trained on a classification task is used to generate CAMs. Second, the CAMs go through a simple post-processing step to remove noise before connected component analysis, which gives the final annotation proposals. We generate the CAMs for all images as rejecting false positives only takes a negligible amount of time compared to polygon annotation and it forces the human annotators to check all the images for false positives/negatives. Optionally, most false negatives could easily be filtered out by applying a classifier to image patches. An example of the initial CAMs and the post-processed output is shown in Figure~\ref{fig:gt_overlay_examples}.

\paragraph{Model Architecture.}

Similarly to \cite{cam_unet}, we adopt U-net~\cite{unet}, a segmentation architecture which aggregates features from different levels of the encoder at different resolutions. Instead of only using the pre-trained encoder for classification, we add the classification head directly on top of the decoder. This approach brings the advantage of having weights pre-trained on the target data as an initialization for subsequent fully supervised learning and increases the resolution of the final layer CAMs.

\begin{figure}[ht]
    \includegraphics[keepaspectratio,width=0.95\columnwidth]{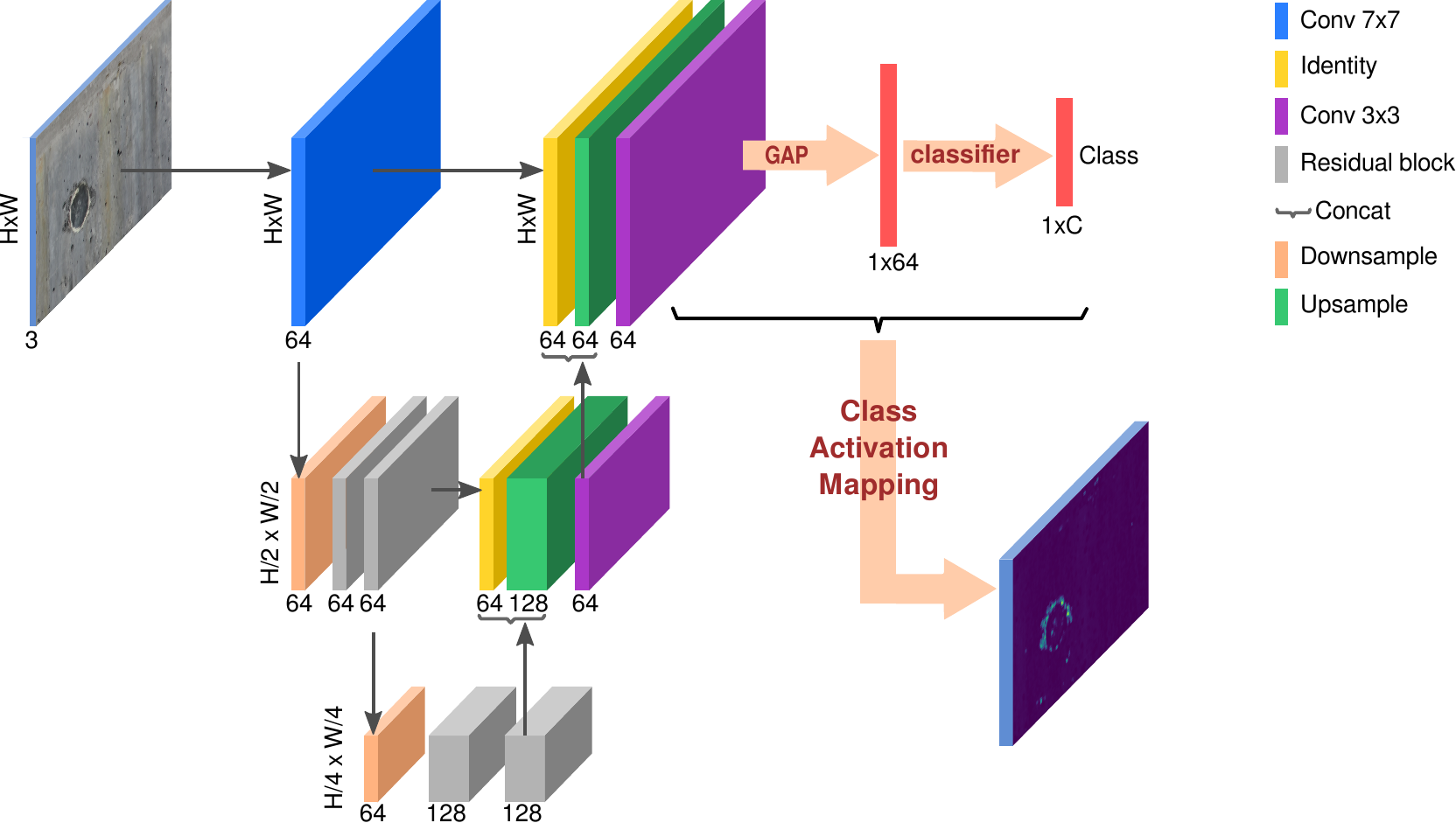}
    \caption{Our U-net-based classifier for assisted labeling via explainability. Using a U-net architecture as the feature extractor of the classifier trained on (weak) classification labels allows us to generate CAMs with the same resolution as the input images using Grad-CAM, a standard gradient-based explainability method. The obtained high-resolution CAMs are then refined using anti-adversarial climbing (\mbox{AdvCAM}), post-processed, and used as proposal segmentation masks that can be further refined by annotators in a standard annotation tool.}
    \label{fig:architecture}
\end{figure}

To further improve the resolution of the CAMs, we only build the U-net on top of the first two blocks of a Resnet34 encoder, avoiding resolution degradation from further downsampling. We also set the stride to 1 instead of the original 2 in the first convolutional layer, before the residual blocks. The model reduction does not lead to any classification performance degradation for the target application, however, Resnet34 produced better quality attribution masks then Resnet18.
An overview of the architecture can be found in Figure~\ref{fig:architecture} and examples of CAMs with a different number of downsampling layers in Figure~\ref{fig:down_cams}.

{
\setlength{\tabcolsep}{1pt}

\begin{figure}[htb]
    \centering
        \begin{tabular}{cccc}
             \includegraphics[keepaspectratio, width=0.24\linewidth]{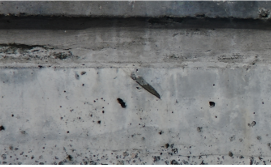} & 
             \includegraphics[keepaspectratio, width=0.24\linewidth]{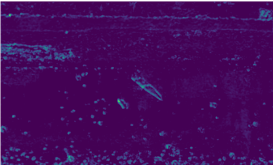} &
             \includegraphics[keepaspectratio, width=0.24\linewidth]{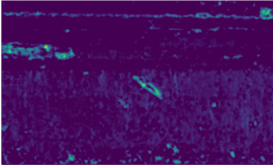} &
             \includegraphics[keepaspectratio, width=0.24\linewidth]{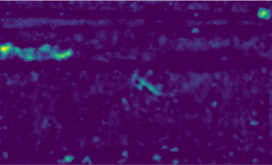} \\
             image & 1 down. layer & 2 down. layer & 3 down. layer  \\
        \end{tabular}
    \caption{Sparsity vs. CAM resolution trade-off. `Spalling' CAMs of networks with one, two and three downsampling layers are shown. The less downsampling layers, the sparser and with better resolution the maps are.}
    \label{fig:down_cams}
\end{figure}

}

\paragraph{Masks as Attribution Maps.}
GradCAM~\cite{selvaraju2017grad} is a gradient-based attribution method used to explain the predictions of a deep neural classifier by localizing class-discriminative regions of an image. It can be used for any layer but in the context of weakly supervised segmentation, the attribution maps of the final layer are typically used. When the last convolutional layer is followed by global average pooling (GAP) and a linear layer as classifier, these maps are commonly referred to as class activation maps (CAMs).

\paragraph{Refinement with Adversarial Climbing.}
The attribution maps obtained with GradCAM typically only reflect the most discriminative regions of an object. Recent methods aim to mitigate this in different ways, by extending the attributed regions of the target object. In our framework, we adopt AdvCAM ~\cite{advcam}, which produces the CAMs on top of images obtained by iterative anti-adversarial manipulation, maximizing the predicted score of a given class while regularizing already salient regions. We observe that for the civil infrastructure domain, a much smaller number of iterations (2) is needed as opposed to the original work (27) to output CAMs with the entire (or almost entire) object region covered. Ideally, as the optimal number of iterations may vary per image, the number of iterations is adjustable by the user in an annotation tool.

The idea of AdvCAM is inspired by that of an non-targeted gradient adversarial attack where a small perturbation is applied to an image $x$ so that the perturbed image $x'$ confuses the classifier into predicting a different class:

\begin{equation}
    x^{\prime}=x-\xi \nabla_{x} \mathrm{NN}(x).
\end{equation}

In AdvCAM, instead of minimizing the score of the target class $c$, the goal is to maximize it by applying: 

\begin{equation}
    x^{\prime}=x+\xi \nabla_{x} y_{c}
\end{equation}

\noindent where $y_c$ is the logit of the target class.

This is referred to as anti-adversarial climbing and the procedure is iterative. Two forms of regularization are also introduced: i) the logits of the other classes are restricted to avoid increase in score for objects of classes close to the target class, and ii) attributions of already salient regions are restricted so that new regions are discovered. 

Finally, the CAMs obtained from the adversarially-manipulated images are summed over all iterations and normalized.
For more details, please refer to AdvCAM ~\cite{advcam}.

\paragraph{Post-processing.}
However, after the previous refinement step, the resulting CAMs contain noise, especially for images with highly structured background. Adversarial climbing typically further increases the amount of noise. Single threshold binarization either includes the noise for lower values, or defect parts are suppressed alongside the noise for higher values.
Due to the increased resolution, the resulting activation maps are also sparser,
sometimes leading to a defect split into multiple parts, especially for very thin cracks.
We add two fast and simple post-processing steps after binarization with a low threshold value, $\theta = 0.1 $,
to address these issues. First, morphological closure is applied to counter the sparsity.
Second, connected components are retrieved from the mask and all regions with
an area below a threshold are filtered out. These steps effectively remove the majority
of the noise while retaining the defect regions. The threshold value $\theta$ was selected according to the best performance on the validation set. The closure filter and minimal component area were selected based on observation of qualitative results.  An example illustrating these steps and the post-processed result is shown in Figure~\ref{fig:postproc}.

 {
\setlength{\tabcolsep}{1pt}

\begin{figure}[htb]
    \centering
        \begin{tabular}{ccccc}
             \includegraphics[keepaspectratio, width=0.20\linewidth]{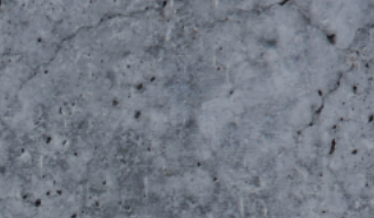} & 
             \includegraphics[keepaspectratio, width=0.20\linewidth]{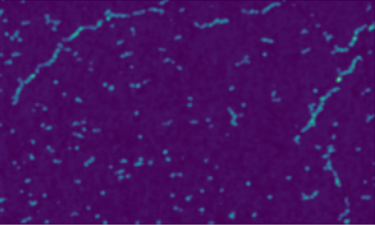} &
             \includegraphics[keepaspectratio, width=0.20\linewidth]{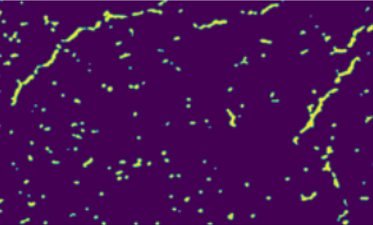} &
             \includegraphics[keepaspectratio, width=0.20\linewidth]{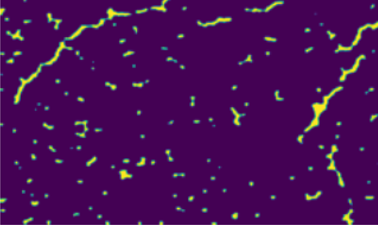} & 
             \includegraphics[keepaspectratio,  width=0.20\linewidth]{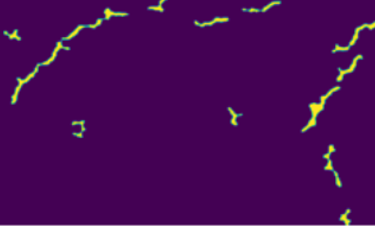} \\
             Original image & CAM & Thresholding & Closure & Filtering \\
        \end{tabular}
    \caption{Postprocessing for noise removal from CAMs. The figure shows the essential post-processing step from the initial CAMs to the final proposed annotation. The CAMs are first binarized with a low threshold (in this case $\theta = 0.1$). We then apply morphological closure to the binary map, followed by filtering of connected components by area to remove small clusters of pixels.}
    \label{fig:postproc}
\end{figure}

}

\subsection{Interactive Aspects}

The quality of the automatically generated annotations varies considerably.
We therefore treat them as proposals to be screened by human annotators in a selection phase, before training a segmentation model on them.
The generated annotations are split into three groups:
\begin{itemize}
\item \textit{Accept} -- the proposed annotation covers the predicted defect and no modifications to the defect mask are needed;
\item \textit{Modify and accept} -- the proposed annotation covers the predicted defect, but the mask needs refinement;
\item \textit{Reject} -- the proposed annotation does not contain the predicted defect, as it is a false positive.
Furthermore, the annotator checks if some defects have been missed by the model (false negatives), in which case a manual annotation process should take place.

\end{itemize}

\paragraph{Annotation modification.}

Depending on the proposed defect mask and the capabilities of the annotation tool used, the modification can take multiple forms.
The most common one is the removal of false positive patches of ``noise'' from the neighbourhood of the defect, which can be done by simply clicking on them to erase them. Another commonly needed modification is removing or adding a small part of the defect, which requires moving/adding/removing polygon points, or with a brush tool. The interaction time can be further reduced by more sophisticated operations, for example, merging and splitting components, mask erosion and dilation, if the annotation tool in use allows for such operations.

\paragraph{Interaction logging.}

The interaction of the annotator with the proposals provides valuable information that can be exploited if logged. For instance, flagged false positive and false negative regions could be used to extend the weakly supervised training set, which consequently would improve the classifier and the subsequent generated proposals.
The time spent on the modification of a proposal until it is accepted can also be used as a proxy for the sample difficulty, allowing for more efficient training strategies.
This working modality of our proposal could be used in the future as part of a labeling pipeline combining active learning pipeline and weak supervision.


\paragraph{Demonstration.}

\begin{figure*}[t]
    \centering
        \begin{tabular}{ccc}
             \includegraphics[keepaspectratio, height=2.65cm]{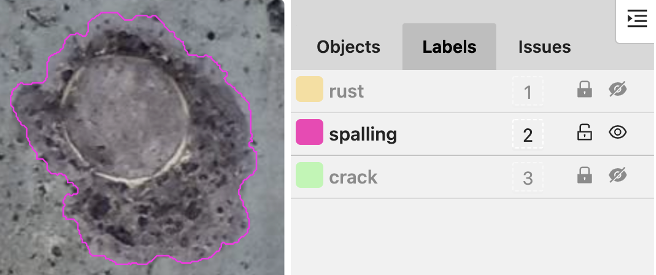} & 
             \includegraphics[keepaspectratio, height=2.65cm]{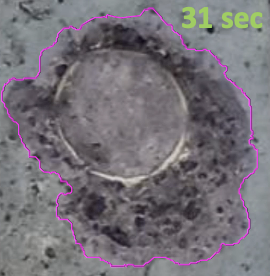} &
             \includegraphics[keepaspectratio, height=2.65cm]{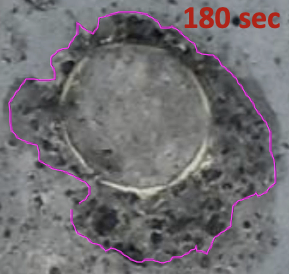} \\
             Proposed annotation & Refined annotation & Manual annotation \\
        \end{tabular}
    \caption{Difference in labeling time between manually refining a proposed annotation obtain with our method vs. manual annotation from scratch. The proposed annotation obtained through our method for a spalling defect is loaded in CVAT (left). It is manually refined by hand by a human annotator in 31 seconds (center). On the other hand, manual annotation from scratch of the same defect takes $\sim$6 times longer (180 seconds) than refining the proposed annotation (right).}
    \label{fig:cvat}
\end{figure*}

We show user interaction in CVAT, an open source annotation tool extensively used in various domains, including civil infrastructure. There, experts are able to visualize, accept, modify and reject the proposed annotations resulting from our framework. In Figure~\ref{fig:cvat}, refining an annotation proposal for a spalling defect takes 31s, as opposed to manually generating the mask, which takes 180s. Additionally, the manual annotation is less fine-grained (i.e., fewer polygon points) than the proposal and reaching the same level of detail manually would extend well beyond 180s. The example shown here is pessimistic as the proposed and refined masks are very close and it is not clear edits were necessary. However, the refinement took only 16 \% of the full annotation time.

\section{Evaluation}
\subsection{Data preparation}
The defect dataset consists of 732 high resolution ($5\text{K} \times 5\text{K}$ pixels) images.
The classification dataset is generated by sampling 5 crops of size $320 \times 320 $ pixels around each positive (corresponding to a defect) user-annotated pixel. Given the extreme class-imbalance, where the least number of instances per class is 675 and the largest number of instances is 21,787 (see Table \ref{tab:dataset}), we create a separate binary classification (defect/no defect) dataset for each defect type. Since there is significantly more negative pixel annotations, crops of negative class are sampled uniformly from each annotation so that there is the same number of positive and negative samples in each of the datasets. Performance of classifiers trained on multiclass datasets created by over/undersampling was inferior.
\begin{table*}[]
    \setlength{\tabcolsep}{8pt}
    \centering
        \begin{tabular}{l|rrrrr}
        defect &  crack &  spalling &  rust \\
        instances & 21,787 & 675 & 1,078 \\
        \end{tabular}
    \caption{Number of instances of each class in the dataset (highly imbalanced).}
    \label{tab:dataset}
\end{table*}

\subsection{Classifier training}
The models share the same architecture and were implemented in the PyTorch framework. Each model was trained for 6 hours on 2 Nvidia A100 GPUs with the batch size of 32 and the AdamW \cite{loshchilov2017decoupled} optimizer (learning rate 1e-4, weight decay 1e-2, all other parameters were kept at PyTorch default values). The best checkpoint was selected according to the highest f-measure on the validation set. 

\subsection{Estimation of time saved}
\begin{figure*}[t]
    \centering
        \begin{tabular}{ccc}
            50-75 \% & 75-95 \% & 95-100 \% \\
            \includegraphics[height=1.7cm, width=0.27\linewidth]{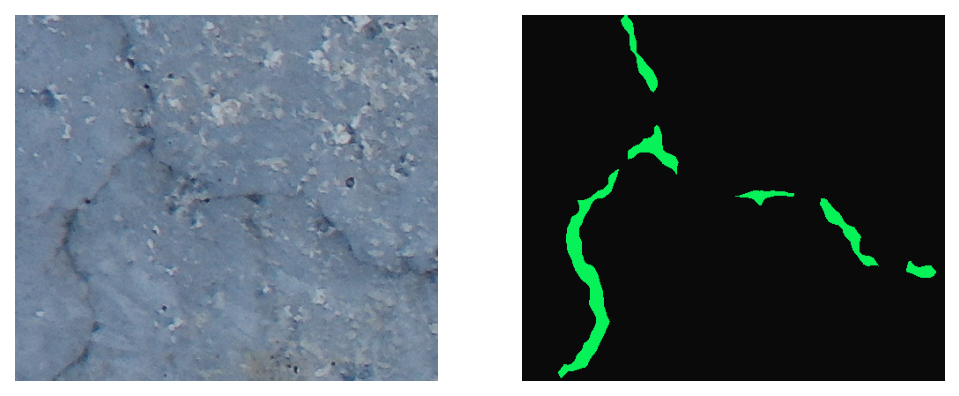} & \includegraphics[height=1.7cm, width=0.27\linewidth]{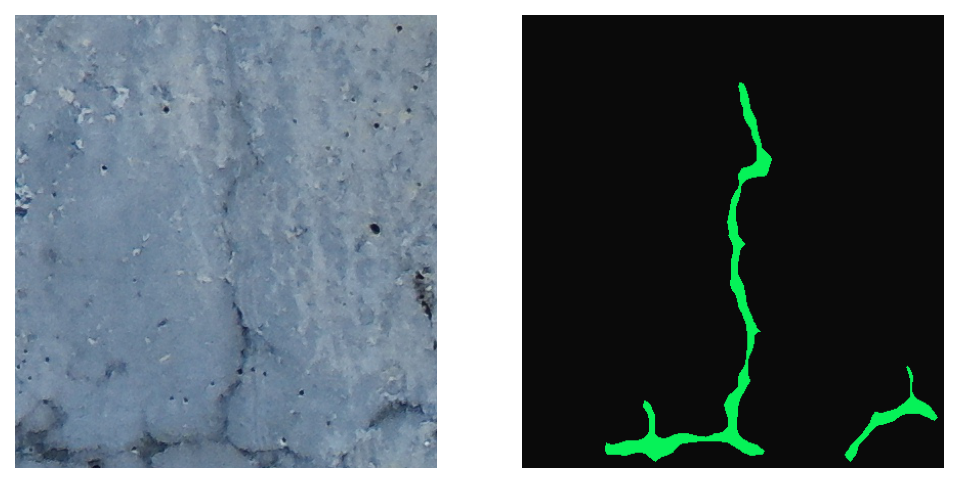} &
            \includegraphics[keepaspectratio, width=0.27\linewidth]{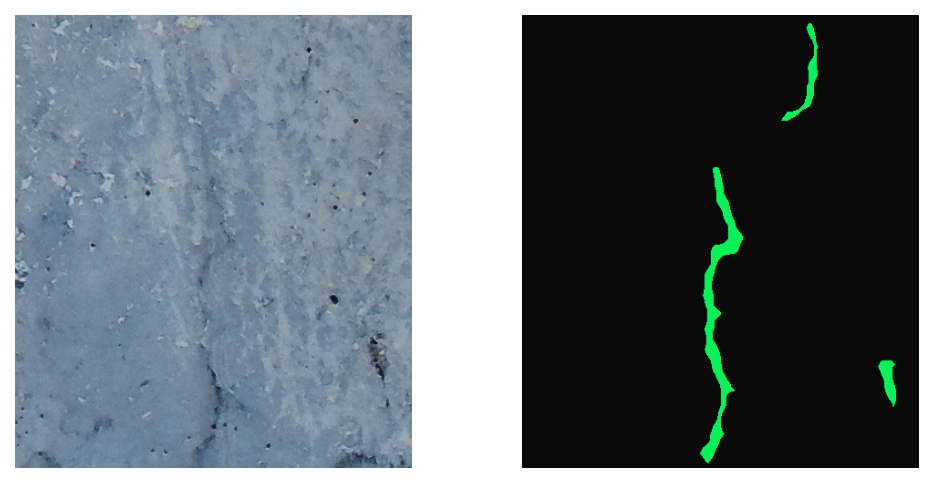} \\
            
            \includegraphics[keepaspectratio, width=0.27\linewidth]{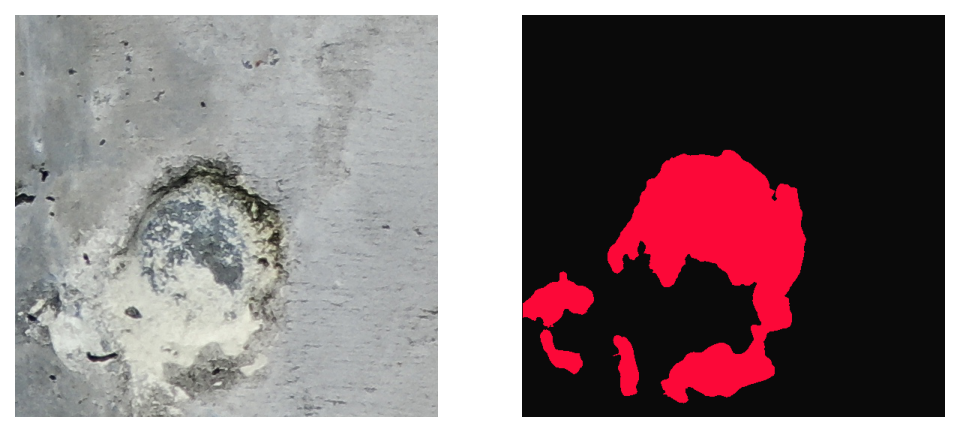} &
            \includegraphics[height=1.5cm, width=0.27\linewidth]{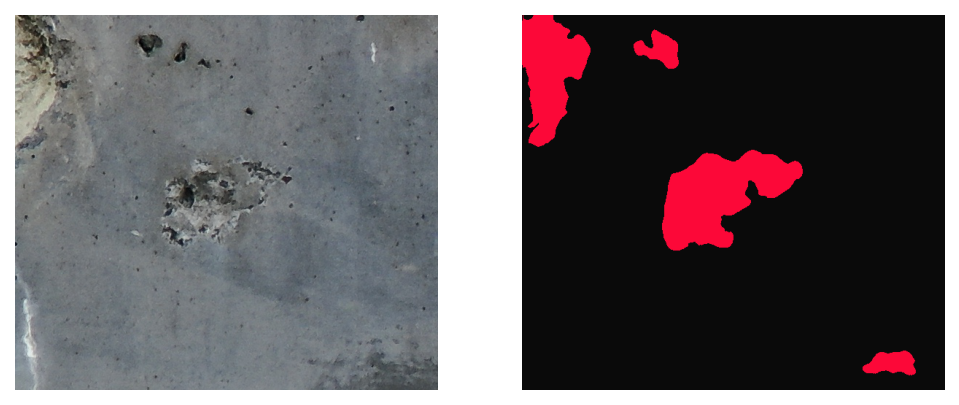} &
            \includegraphics[height=1.5cm, width=0.27\linewidth]{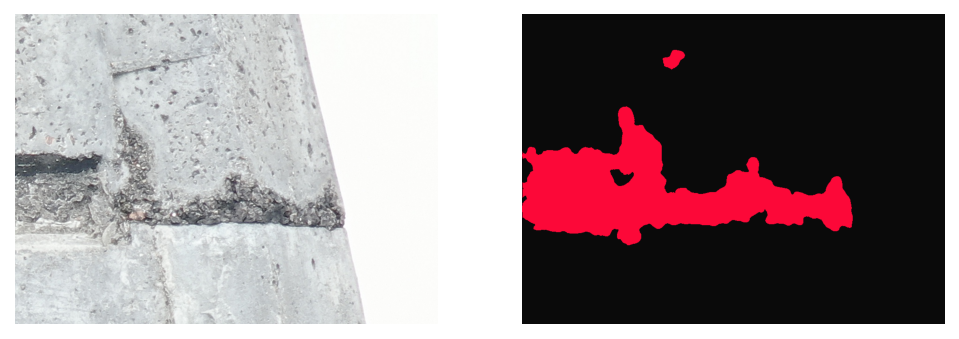}\\
            
            \includegraphics[keepaspectratio, width=0.27\linewidth, trim={0 2.5cm 8.5cm 2.5cm},clip]{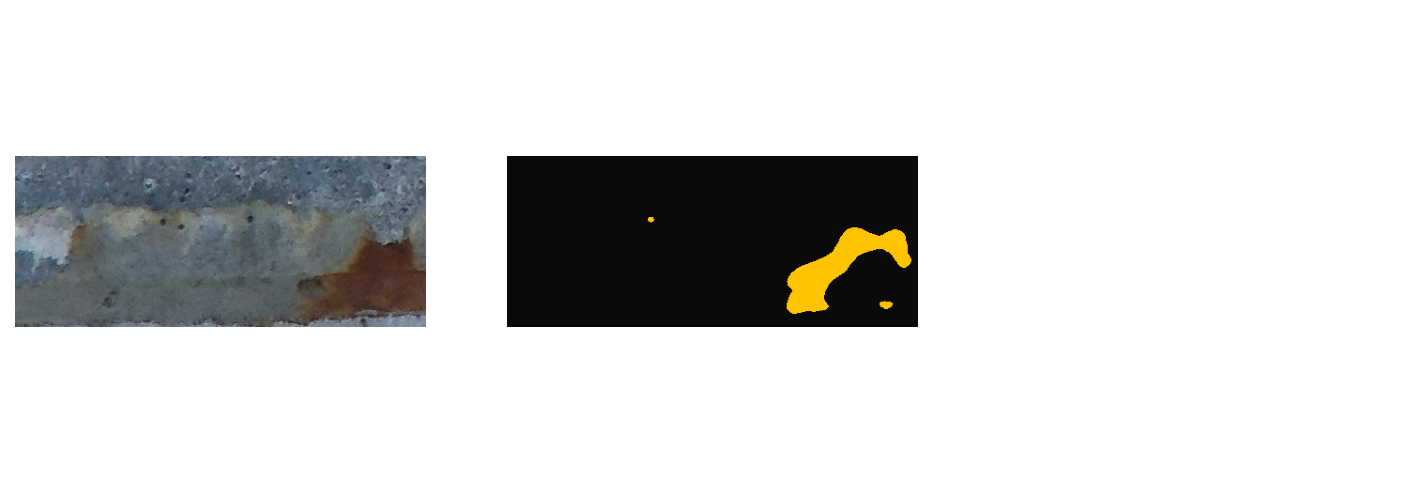} &
            \includegraphics[keepaspectratio, width=0.27\linewidth, trim={0 2.5cm 8.5cm 2.5cm},clip]{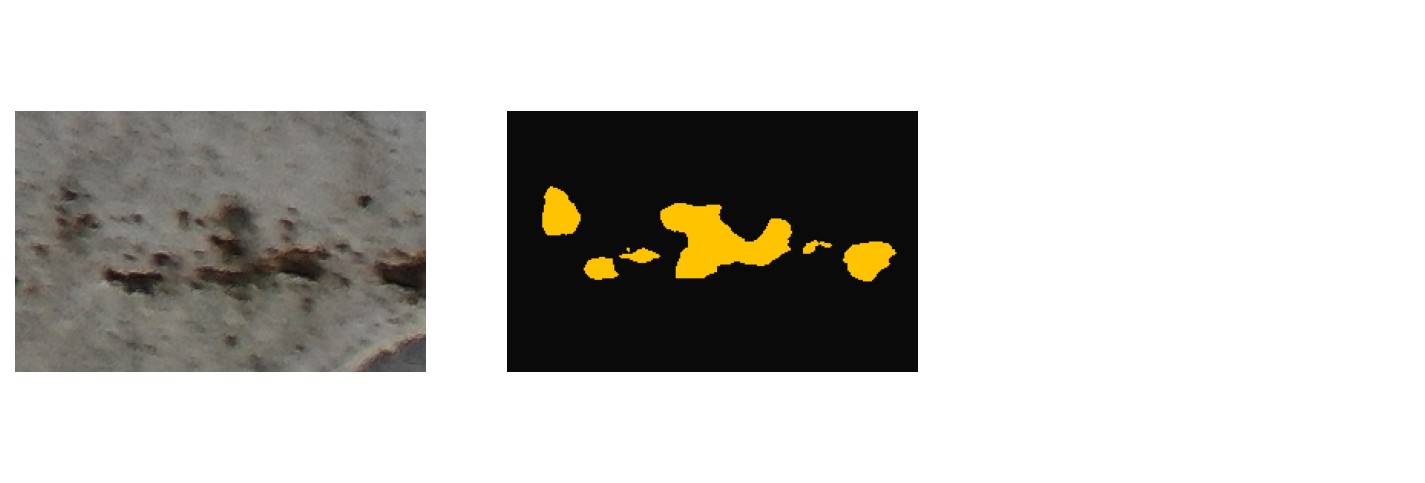} &
            \includegraphics[keepaspectratio, width=0.27\linewidth, trim={0 2.5cm 8.5cm 2.5cm},clip]{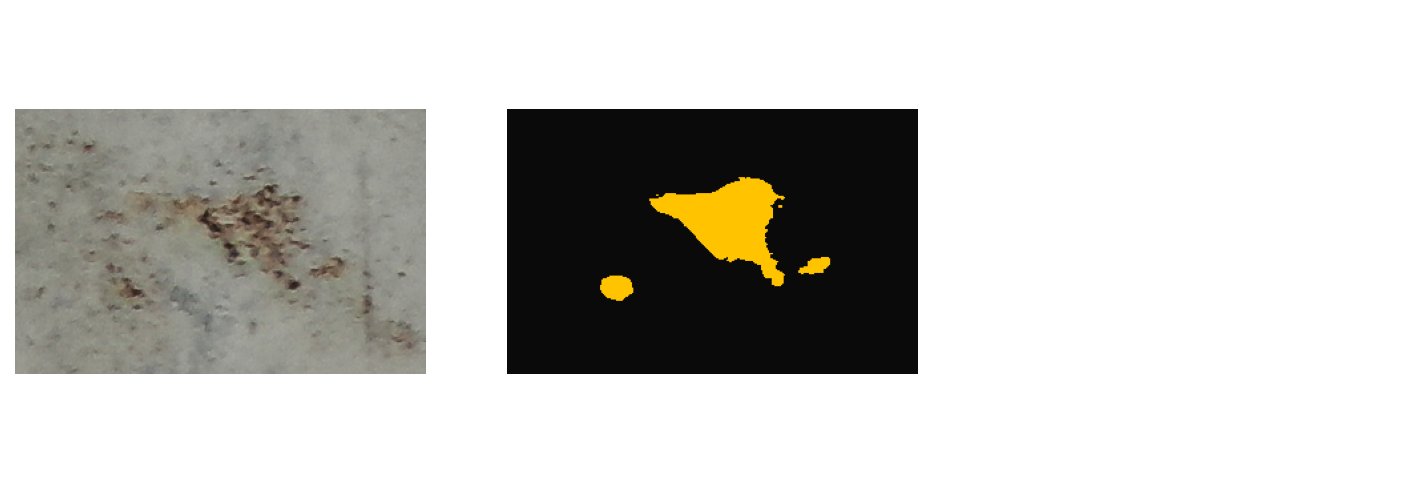} \\
        \end{tabular}
    \caption{Examples of instances according to the annotation time saved (in \%). The first, second and last row show instances of cracks, spalling and rust, respectively.}
    \label{fig:time_saved_examples}
\end{figure*}

To quickly estimate the annotation time reduction due to our weakly supervised model, we used the following procedure where users only estimate the percentage of time saved, as opposed to actually annotating the data. 
We first assumed that the user has at their disposal standard annotation tools such as brush and erasure, as well as the possibility to apply morphological operations such as dilation and erosion.
We then split the test set instances (i.e.\ the connected components of the segmentation ground truth masks in the test patches, which are $265$ in total) into 3 groups according to the estimated percentage of the time saved annotating the instances when using the output of our method as an initial annotation proposal within such a standard annotation tool.
Specifically, group $\text{G}_{95}$ are the instances for which, by visual inspection, we estimated modifying the CAM via the annotation tool provided a time saving above $95\%$ over annotating the instance from scratch.
Analogously, groups $\text{G}_{75}$ and $\text{G}_{50}$ are groups of instances where we estimated a time saving above $75\%$ (but below $95\%$) and above $50\%$ (but below $75\%$), respectively. Examples of instances from each group for all defects are shown in Figure \ref{fig:time_saved_examples}.

The ratio between the total time saved and the time needed to annotate was then simply estimated from the number of instances in each of the groups of connected components as:
%
\begin{equation}
     \mbox{Relative time saving} = \frac{1}{N} \sum_{i \in \{95, 75, 50 \} }{ \frac{i}{100}|\text{G}_i| },
\end{equation}
where $|\text{G}_i| $ is the number of instances in the group.

This formula allowed us to estimate an average reduction of 52\% in annotation time.
This is broken down as follows for different types of defects: 57\% for cracks, 58\% for spalling and 40\% for rust.
Detailed results of this estimation procedure are reported in Table \ref{tab:time_saved}. The relatively low time saving on rust can be explained by the high sensitivity of the annotation proposal shape to the value of the threshold applied to the CAMs, meaning that each component would require fine manual tuning of the threshold within the annotation tool.

An important note on the limitations of this time estimation procedure is that on one hand, it does not take into account the instances that were missed by our method, meaning that in practice we assumed  recall of the defects.
On the other, the time estimation is very conservative. Less than 50 \% time reduction is considered as 0 and a lower bound is used for the rest of the intervals. The final result of 52\% annotation time saved should thus be considered as a very conservative lower bound and we plan to conduct a more detailed evaluation in the future.

\begin{table*}[]
    \setlength{\tabcolsep}{8pt}
    \centering
        \begin{tabular}{lrrrrr}
        \toprule
        {} &  instance count &  95 &  75 &  50 &  time saved (\%)\\
        \midrule
        crack       &             111 &  19 &  40 &  30 &          57 \\
        spalling    &              65 &  17 &  19 &  14 &          58 \\
        rust        &              89 &  22 &  14 &   9 &          40 \\
        all defects &             265 &  58 &  73 &  53 &          51 \\
        \bottomrule
        \end{tabular}
    \caption{Estimated annotation time saved by our method in percentage over annotating the defect instances from scratch on a test set of 265 instances. Relative time saving is shown for each type of defect separately and in total averaged over all defects. The number of instances where the percentage of time saved $t \in [95, 100]$, $t \in [75, 95)$ and $t \in [50, 75)$ is also reported. Anything less than $50$ is considered as no time saved on the instance.}
    \label{tab:time_saved}
\end{table*}

\section{Conclusions}

We explored the use of weak labels from human annotators as a means to reduce the labeling time for a segmentation task in visual inspection, an application domain where the time of specialized annotators is particularly costly.
The advantage of weak labels is that they are cheap to obtain because they require minimal interaction with the annotator.
In our proposed approach, weak labels are used to train a classifier in order to generate proposals for segmentation masks by means of an explainability attribution method, followed by iterative adversarial climbing.
Domain experts can then correct the proposed masks where needed by integrating this workflow in standard annotation tools like CVAT.
Moreover, proposal segmentation masks can be used as pseudo-labels for unlabeled images, which can subsequently be employed to train supervised segmentation models, as well as to diagnose issues with ground truth labels from previous annotation campaigns.

\subsubsection*{Acknowledgement}
This research was supported by Czech Technical University
student grant SGS20/171/OHK3/3T/13.
We would like to thank Finn Bormlund and Svend Gjerding from Sund\&Bælt for the collection and annotation of the image data.

\clearpage
%
%
\bibliographystyle{splncs04}
\bibliography{egbib}
\end{document}